# Bayesian Optimization that Limits Search Region to Lower Dimensions Utilizing Local GPR

Yasunori Taguchi
*Corporate R&D Center*
*Toshiba Corporation*
Kanagawa, Japan
yasunori.taguchi@toshiba.co.jp

Hiro Gangi
*Corporate R&D Center*
*Toshiba Corporation*
Kanagawa, Japan
hiro1.gangi@toshiba.co.jp

*Abstract*— Optimization of product and system characteristics is required in many fields, including design and control. Bayesian optimization (BO) is often used when there are high observing costs, because BO theoretically guarantees an upper bound on regret. However, computational costs increase exponentially with the number of parameters to be optimized, decreasing search efficiency. We propose a BO that limits the search region to lower dimensions and utilizes local Gaussian process regression (LGPR) to scale the BO to higher dimensions. LGPR treats the low-dimensional search region as "local," improving prediction accuracies there. The LGPR model is trained on a local subset of data specific to that region. This improves prediction accuracy and search efficiency and reduces the time complexity of matrix inversion in the Gaussian process regression. In evaluations with 20D Ackley and Rosenbrock functions, search efficiencies are equal to or higher than those of the compared methods, improved by about 69% and 40% from the case without LGPR. We apply our method to an automatic design task for a power semiconductor device. We successfully reduce the specific on-resistance to 25% better than a conventional method and 3.4% better than without LGPR.

## I. Introduction

Searches for parameters with favorable characteristics are often performed for the design of devices and materials, control of systems and robots, and hyperparameter tuning for machine learning models. This involves the repeated determination of parameter values for observing characteristic values (a point in the parameter space) and observing characteristic values at that point by simulations or experiments. Such observations are often costly in terms of time, money, and human labor, so it is desirable to obtain optimal parameters from a small number of observations. In many black-box optimization algorithms [1], Bayesian optimization (BO) [2]-[4], which guarantees an upper bound on regret, is often used to minimize those costs.

In BO, Gaussian process regression (GPR) [5] is typically used as a surrogate model for the objective function, for which neither the value at each point nor the derivative is known. A parameter value that maximizes the acquisition function obtained from this surrogate model is determined as the next observation query point. This approach has been successful in many areas [6]-[11]. However, the dimensionality of the space increases with the number of parameters, exponentially increasing the complexity of full-domain observations, surrogate model computations, and acquisition function maximizations [12], [13]. BO is thus difficult to apply to high-dimensional optimization problems.

Inspired by LineBO [14], which limits search region to one-dimension, we propose a BO that trains GPR model locally around lower-dimensional search region to scale the BO to higher dimensions. We call such BO with low-dimensionally limited search spaces BOLD. BOLD includes, but is not limited to, LineBO. To reduce the time complexity of matrix inversion and improve the accuracy of GPR predictions in a low-dimensional search region that requires calculations of the acquisition function, a local subset of data (LSoD) specific to that region is extracted and the local GPR (LGPR) model is trained. We call BOLD utilizing local GPR BOLDUC. The following presents the results of evaluation experiments, confirming the effectiveness of our method.

### A. Contributions

The main contributions of this paper are as follows:

- We propose BOLDUC.

- We introduce the concept of the contribution of observed points for LGPR prediction in low-dimensional search regions. We present three LSoD extraction strategies based on the proposed method, training the LGPR model on the LSoD. To our knowledge, no prior works on LGPR limit prediction targets to points in a low-dimensional search region.

- BOLDUC inherits the theoretical guarantees in [14] when LineBO is adopted as BOLD.

- Similar to the global subset of data (SoD) [15], we reduce the time complexity of matrix inversion in GPR from $\mathcal{O}(N^3)$ to $\mathcal{O}(M^3)$ ($M < N$). We can overcome upper bounds on the number of observations, since $M$ is set sufficiently small according to the low dimensionality of the local search region.

- Optimization tasks involving two benchmark functions and an automatic design task for a power semiconductor device show improved search efficiency as compared to BOLD without LGPR. The latter task in particular demonstrated the automatic acquisition of excellent specific on-resistance ($R_{ON}A$) values.

### B. Related work

In prior works that scale BO to higher dimensions, ADD-GP-UCB [16] and its generalization [17] assume additivity in the structure of the objective function and optimize in low-dimension groups. REMBO [18], HesBO [19], and ALEBO [20] assume a low effective dimensionality and embed a low-dimensional space into the high-dimensional space. These methods are effective when there are many redundant parameters or parameters that behave like noise and have little effect on the objective function's value. In DropoutBO [21], GPR, acquisition function maximization, and solution searches are performed in low-dimensional spaces defined by the coordinate axes selected through dropout. Observation point values in dropped-out axes are determined according to three types of fill-in strategies. LineBO [14] limits the search region to one-dimensional spaces. It can be integrated with SafeOpt [22]. EGP [23] adopts a strategy of gradually decreasing the length scale of the kernel function to tackle the flattening problem of the acquisition function in high dimensions. This gradually shifts from global to local approximation. TuRBO [24] sets local hyperrectangles as trust regions in which to search for the solution. Although it is similar to the proposed method in that it uses LGPR, the local area differs from ours. BAxUS [25] extends TuRBO by using the embedding technique.

Time complexity reduction methods for GPR can be classified into two types: global and local approximations [26]. The simplest method for global approximation is SoD [15]. This reduces the time complexity of GPR to $\mathcal{O}(M^3)$ by extracting $M$ of the observed $N$ points and obtaining a small kernel matrix $K_{M\times M}$. Sparse GPR [27]-[29] constructs the kernel matrix with $L\ (< N)$ inducing points representing all data to maintain the information of vanishing points in SoD as much as possible, reducing the time complexity to $\mathcal{O}(NL^2 + L^3)$. Local approximation divides the entire domain into subregions and a LGPR model corresponding to a subregion based on the $M$ points belonging to that subregion [30]. This reduces the time complexity in each region to $\mathcal{O}(M^3)$ ($M < N$). When predetermined points are targeted for prediction, local regions specialized for them are set [31], [32].

*C. Outline*

This paper is organized as follows: Sect.II starts with a problem statement of black-box optimization and prepares overviews of GPR and standard BO. In Sect. III, BOLDUC is introduced. Experimental evaluations are given in Sect. IV. Conclusions are drawn in Sect. V.

## II. PROBLEM STATEMENT AND PRELIMINARIES

Let $\chi \subset \mathbb{R}^D$ be a domain of parameter vector $\mathbf{x} = (x_1, x_2, \ldots, x_D)^\top$ and $f: \chi \to \mathbb{R}$ the unknown objective function, where $\cdot^\top$ denotes the transpose of a vector or matrix. We wish to find

$$\mathbf{x}_{opt} = \arg\min_{\mathbf{x}\in\chi} f(\mathbf{x}),$$

considering that the optimized parameter value $\hat{\mathbf{x}}$ is better when the simple regret

$$r(\hat{\mathbf{x}}) = f(\hat{\mathbf{x}}) - f(\mathbf{x}_{opt})$$

is smaller.

*A. Gaussian process regression*

In GPR, a function $f$ is regressed from observation data

$$\mathcal{D}_N = \{(\mathbf{x}_n, y_n)\}_{n=1,2,\ldots,N},$$

where $\mathbf{x}_n$ represents the $n$-th observed point and $y_n$ is observed by the following equation with additive noise $\varepsilon_n$:

$$y_n = f(\mathbf{x}_n) + \varepsilon_n.$$

The predicted mean $\mu(\mathbf{x}|\mathcal{D}_N)$ and variance $v(\mathbf{x}|\mathcal{D}_N)$ of the function value at an unobserved point $\mathbf{x}$ by GPR are expressed as

$$\mu(\mathbf{x}|\mathcal{D}_N) = \mathbf{k}^\top (K_{N\times N} + \sigma^2 I_N)^{-1} \mathbf{y}_{1:N}, \quad (1)$$
$$v(\mathbf{x}|\mathcal{D}_N) = k(\mathbf{x},\mathbf{x}) - \mathbf{k}^\top (K_{N\times N} + \sigma^2 I_N)^{-1} \mathbf{k},$$

---

**Algorithm 1** Standard BO

1: Initialize data: $\mathcal{D}_{N_0} = \{(\mathbf{x}_n, y_n)\}_{n=1,2,\ldots,N_0}$.
2: **for** $t = N_0, N_0 + 1, \ldots, N - 1$ **do**
3:   Estimate kernel hyperparameter $\boldsymbol{\theta}(\mathcal{D}_t)$ and train global GPR model $\mathcal{M}(\mathcal{D}_t)$.
4:   Suggest $\mathbf{x}_{t+1} = \arg\max_{\mathbf{x}\in\chi} \alpha(\mathbf{x}|\mathcal{D}_t)$ using $\mathcal{M}(\mathcal{D}_t)$.
5:   Observe $y_{t+1} = f(\mathbf{x}_{t+1}) + \epsilon_{t+1}$.
6:   Augment the data $\mathcal{D}_{t+1} = \mathcal{D}_t \cup \{(\mathbf{x}_{t+1}, y_{t+1})\}$.
7: **end for**
8: Output $\hat{\mathbf{x}}_N = \mathbf{x}_b$, where $b = \arg\min_{n=1,2,\ldots,N} y_n$.

---

where $\mathbf{k} = \big(k(\mathbf{x},\mathbf{x}_1), k(\mathbf{x},\mathbf{x}_2), \cdots, k(\mathbf{x},\mathbf{x}_N)\big)^\top$, the $(i,j)$ component of $K_{N\times N}$ is $k(\mathbf{x}_i, \mathbf{x}_j)$, $\mathbf{y}_{1:N} = (y_1, y_2, \ldots, y_N)^\top$, $k(\cdot,\cdot)$ is the kernel function representing the similarity between two points, $\sigma$ is the standard deviation of the noise component, $I_N$ is the $N \times N$ identity matrix, and $\cdot^{-1}$ denotes the inverse of a matrix. Both $\mu$ and $v$ depend on the hyperparameter $\boldsymbol{\theta}$ of the kernel function. $\boldsymbol{\theta}$ is estimated from $\mathcal{D}_N$ by maximizing the marginal likelihood $p(\mathbf{y}_{1:N}|\{\mathbf{x}_n\}_{n=1,2,\ldots,N}, \boldsymbol{\theta})$ whose logarithm is proportional to

$$-\log|K_{N\times N} + \sigma^2 I_N| - \mathbf{y}_{1:N}^\top (K_{N\times N} + \sigma^2 I_N)^{-1} \mathbf{y}_{1:N} \\ +\text{const.} \quad (2)$$

Since $\boldsymbol{\theta}$ depends on $\mathcal{D}_N$, $\boldsymbol{\theta}$ is written as $\boldsymbol{\theta}(\mathcal{D}_N)$. The $\mathcal{O}(N^3)$ computation for the matrix inversion and determinant can become major bottlenecks as $N$ grows.

*B. Standard bayesian optimization*

Algorithm 1 shows the standard BO [2]-[4] procedure. At line 1, the data $\mathcal{D}_{N_0} = \{(\mathbf{x}_n, y_n)\}_{n=1,2,\ldots,N_0}$ is prepared. At line 3, kernel hyperparameter $\boldsymbol{\theta}(\mathcal{D}_t)$ is estimated using (2) with $N$ replaced by $|\mathcal{D}_t|\ (=t)$ and global GPR model $\mathcal{M}(\mathcal{D}_t)$ is learned from $\mathcal{D}_t$. At line 4, the acquisition function $\alpha$ determined from $\mathcal{M}(\mathcal{D}_t)$ finds the maximum point in the domain $\chi$ and defines it as the next observed query point

$$\mathbf{x}_{t+1} = \arg\max_{\mathbf{x}\in\chi} \alpha(\mathbf{x}|\mathcal{D}_t).$$

For example, the lower confidence bound (LCB) [3] is used as the acquisition function

$$\alpha(\mathbf{x}|\mathcal{D}_t) = -\left(\mu(\mathbf{x}|\mathcal{D}_t) - \kappa_t\sqrt{v(\mathbf{x}|\mathcal{D}_t)}\right),$$

where $\kappa_t$ is a parameter that adjusts the balance between exploitation and exploration. At line 5, $y_{t+1}$ corresponding to $\mathbf{x}_{t+1}$ is observed. At line 6, the data are augmented. At line 8, the optimized parameter value $\hat{\mathbf{x}}_N$ is obtained.

In this method, the search space expands exponentially with the number of dimensions $D$, increasing the computational complexity and worsening the parameter search efficiency. When the number of observation samples $|\mathcal{D}_t|$ is large, the time complexity $\mathcal{O}(|\mathcal{D}_t|^3)$ (where $|\cdot|$ denotes the number of elements in the set) of GPR may become unignorable compared to the observation time.

## III. BOLDUC

*A. BOLD*

Algorithm 2 shows the BOLD procedure. Lines 3 and 5 differ from the standard BO. At line 3, a low-dimensional affine subspace passing through the best point $\hat{\mathbf{x}}_t = \mathbf{x}_{b_t}$,

**Algorithm 2** BOLD (without LGPR)
1: Initialize data: $\mathcal{D}_{N_0} = \{(\mathbf{x}_n, y_n)\}_{n=1,2,\ldots,N_0}$.
2: **for** $t = N_0, N_0 + 1, \ldots, N - 1$ **do**
3:    Define low-dimensional search space $\mathcal{S}_t$.
4:    Estimate kernel hyperparameter $\boldsymbol{\theta}(\mathcal{D}_t)$ and train global GPR model $\mathcal{M}(\mathcal{D}_t)$.
5:    Suggest $\mathbf{x}_{t+1} = \arg\max_{\mathbf{x} \in (\mathcal{S}_t \cap \chi)} \alpha(\mathbf{x}|\mathcal{D}_t)$ using $\mathcal{M}(\mathcal{D}_t)$.
6:    Observe $y_{t+1} = f(\mathbf{x}_{t+1}) + \epsilon_{t+1}$.
7:    Augment the data $\mathcal{D}_{t+1} = \mathcal{D}_t \cup \{(\mathbf{x}_{t+1}, y_{t+1})\}$.
8: **end for**
9: Output $\hat{\mathbf{x}}_N = \mathbf{x}_b$, where $b = \arg\min_{n=1,2,\ldots,N} y_n$.

where $b_t = \arg\min_{n=1,2,\ldots,t} y_n$, is set as the low-dimensional search space

$$\mathcal{S}_t = \hat{\mathbf{x}}_t + \mathcal{U}_t = \{\hat{\mathbf{x}}_t + \mathbf{u}_t | \mathbf{u}_t \in \mathcal{U}_t\},$$

where $\mathcal{U}_t$ is the $d_t (< D)$-dimensional linear subspace associated with $\mathcal{S}_t$. $\mathcal{S}$ is changed when the suggested queries stagnate at a nearby position or when a predetermined number of values are observed. At line 5, the next query $\mathbf{x}_{t+1}$ is determined from within the search region $(\mathcal{S}_t \cap \chi)$. LineBO [14] is a BOLD with $d_t$ fixed to 1.

Since $y$ is heavily sampled in the low-dimensional region $(\mathcal{S}_t \cap \chi)$, the surrogate model is more accurate in that region. $d_t < D$ makes it easier to maximize the acquisition function. As a result, the search efficiency in high dimensions is improved. However, BOLD has two problems:

- The kernel hyperparameter $\boldsymbol{\theta}(\mathcal{D}_t)$ is optimal for reproducing $\mathcal{D}_t$ in terms of the marginal likelihood. However, $\boldsymbol{\theta}(\mathcal{D}_t)$ is not necessarily suitable for expressing the local structure of objective $f$ in low-dimensional search region $(\mathcal{S}_t \cap \chi)$. Therefore, even if $\boldsymbol{\theta}(\mathcal{D}_t)$ is used, the GPR prediction accuracy in $(\mathcal{S}_t \cap \chi)$ is insufficiently high.

- $\mathcal{D}_t$ is used to train the global GPR model. The time complexity of the inverse matrix is $\mathcal{O}(|\mathcal{D}_t|^3)$, as with the standard BO, which is large.

### B. LGPR based on LSoD

BOLDUC uses LGPR to solve the above two problems simultaneously. BOLD limits calculations of the acquisition function to the low-dimensional local region $(\mathcal{S}_t \cap \chi)$. The LGPR model is constructed with SoD with fewer samples than $\mathcal{D}_t$, excluding samples from observation points far from $(\mathcal{S}_t \cap \chi)$ to improve GPR prediction accuracy within $(\mathcal{S}_t \cap \chi)$ and to reduce the time complexity of GPR. The local domain of $\mathbf{x}$ targeted for prediction differs from that in conventional LGPRs. They are clusters in [30], local hyperrectangles in TuRBO [24], and points in [31], [32], while in ours it is the low-dimensional search region $(\mathcal{S}_t \cap \chi)$.

Equation (1) is also expressed as

$$\mu(\mathbf{x}|\mathcal{D}_t) = \sum_{n=1}^{t} k(\mathbf{x}, \mathbf{x}_n)((K_{t \times t} + \sigma^2 I_t)^{-1} \mathbf{y}_{1:t})_{[n]},$$

where $\cdot_{[n]}$ denotes the $n$-th element of the vector, $((K_{t \times t} + \sigma^2 I_t)^{-1} \mathbf{y}_{1:t})_{[n]}$ is a constant determined from observed data $\mathcal{D}_t$, and the similarity $k(\mathbf{x}, \mathbf{x}_n)$ is a non-negative variable that depends on the unobserved point $\mathbf{x}$. This equation is known as the representer theorem [5]. Because $k(\mathbf{x}, \mathbf{x}_n)$ can be regarded as a weight for the constant $((K_{t \times t} + \sigma^2 I_t)^{-1} \mathbf{y}_{1:t})_{[n]}$, it is defined as follows.

**Definition 1** (Contribution to point). *The similarity $k(\mathbf{x}, \mathbf{x}_n)$ of an observed point $\mathbf{x}_n$ to a point $\mathbf{x}$ is called the contribution of $\mathbf{x}_n$ to the GPR prediction at $\mathbf{x}$.*

To get good prediction accuracy for all points in $(\mathcal{S}_t \cap \chi)$, it is useful to have samples with nonnegligible contributions to at least one point in $(\mathcal{S}_t \cap \chi)$. $\exists \mathbf{x} \in (\mathcal{S}_t \cap \chi), k(\mathbf{x}, \mathbf{x}_n) \geq \lambda_t$ is equivalent to $\max_{\mathbf{x} \in (\mathcal{S}_t \cap \chi)} k(\mathbf{x}, \mathbf{x}_n) \geq \lambda_t$, where $\lambda_t$ is a threshold.

**Definition 2** (Contribution to low-dimensional search region). *The maximum similarity $\max_{\mathbf{x} \in (\mathcal{S}_t \cap \chi)} k(\mathbf{x}, \mathbf{x}_n)$ is called the contribution of an observed point $\mathbf{x}_n$ to the GPR prediction in the low-dimensional search region $(\mathcal{S}_t \cap \chi)$.*

LSoD $\mathcal{E}_t$ is extracted from $\mathcal{D}_t$ samples with nonnegligible contributions to $(\mathcal{S}_t \cap \chi)$ to obtain high prediction accuracy at all points in $(\mathcal{S}_t \cap \chi)$:

$$\mathcal{E}_t = \left\{(\mathbf{x}_n, y_n) \middle| \max_{\mathbf{x} \in (\mathcal{S}_t \cap \chi)} k(\mathbf{x}, \mathbf{x}_n) \geq \lambda_t, n \leq t\right\}. \quad (3)$$

The LGPR model $\mathcal{M}(\mathcal{E}_t)$ is trained on $\mathcal{E}_t$. Thus, LGPR adaptively switches the training data according to $(\mathcal{S}_t \cap \chi)$.

From (3), $\mathcal{E}_t$ depends on the hyperparameter $\boldsymbol{\theta}$ of the kernel $k(\mathbf{x}, \mathbf{x}_n)$. To grasp the global structure of objective $f$ for LSoD extraction, $\boldsymbol{\theta}$ should be estimated from $\mathcal{D}_t$. However, if (2) with $N$ replaced by $|\mathcal{D}_t|$ is used for the estimation, the time complexity cannot be reduced. Hence, as $\boldsymbol{\theta}$ for extracting $\mathcal{E}_t$, we use $\boldsymbol{\theta}(\mathcal{E}_{t-1})$ estimated when the LGPR model $\mathcal{M}(\mathcal{E}_{t-1})$ was constructed at the previous time $t - 1$:

$$\mathcal{E}_t = \left\{(\mathbf{x}_n, y_n) \middle| \max_{\mathbf{x} \in (\mathcal{S}_t \cap \chi)} k(\mathbf{x}, \mathbf{x}_n | \boldsymbol{\theta}(\mathcal{E}_{t-1})) \geq \lambda_t, n \leq t\right\}.$$

When $\mathcal{E}_t$ is extracted for the first time, $\boldsymbol{\theta}(\mathcal{D}_{N_0})$ is estimated from the initial data $\mathcal{D}_{N_0}$ and used instead of $\boldsymbol{\theta}(\mathcal{E}_{t-1})$, since $\boldsymbol{\theta}(\mathcal{E}_{t-1})$ does not exist.

To simplify the calculation, we approximate as follows:

$$\max_{\mathbf{x} \in (\mathcal{S}_t \cap \chi)} k(\mathbf{x}, \mathbf{x}_n) \approx \max_{\mathbf{x} \in \mathcal{S}_t} k(\mathbf{x}, \mathbf{x}_n). \quad (4)$$

**Theorem 1** (Approximation of $\mathcal{E}_t$). *When adopting a kernel function with a higher similarity as the Euclidean distance between two points is shorter, under the approximation expressed in (4), the LSoD $\mathcal{E}_t$ defined in (3) can be rewritten as*

$$\mathcal{E}_t \approx \{(\mathbf{x}_n, y_n) | \|P_{\mathcal{S}_t}(\mathbf{x}_n) - \mathbf{x}_n\| \leq \tau_t, n \leq t\}, \quad (5)$$

*where $\tau_t$ is a threshold and $P_{\mathcal{S}_t}$ is the orthogonal projection operator to the low-dimensional search space $\mathcal{S}_t$:*

$$P_{\mathcal{S}_t}(\mathbf{x}_n) = (I_D - P_{\mathcal{U}_t})\hat{\mathbf{x}}_t + P_{\mathcal{U}_t}\mathbf{x}_n, \quad (6)$$

*where $P_{\mathcal{U}_t}$ represents the orthogonal projection matrix onto the $d_t$-dimensional linear subspace $\mathcal{U}_t$ associated with $\mathcal{S}_t$.*

***Derivation of Theorem 1.*** From the fact that the shorter the Euclidean distance between two points, the higher the similarity between them, and from $\forall \mathbf{x} \in \mathcal{S}_t, \|P_{\mathcal{S}_t}(\mathbf{x}_n) - \mathbf{x}_n\| \leq \|\mathbf{x} - \mathbf{x}_n\|$ (Fig. 1),

$$\max_{\mathbf{x} \in \mathcal{S}_t} k(\mathbf{x}, \mathbf{x}_n) = k(P_{\mathcal{S}_t}(\mathbf{x}_n), \mathbf{x}_n) \quad (7)$$

holds. From (4) and (7), $\max_{\mathbf{x} \in (\mathcal{S}_t \cap \chi)} k(\mathbf{x}, \mathbf{x}_n) \approx k(P_{\mathcal{S}_t}(\mathbf{x}_n), \mathbf{x}_n)$

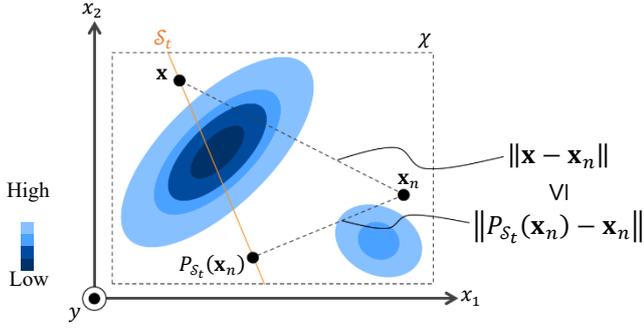

Fig. 1. Comparison between $\|P_{S_t}(\mathbf{x}_n) - \mathbf{x}_n\|$ and $\|\mathbf{x} - \mathbf{x}_n\|$.

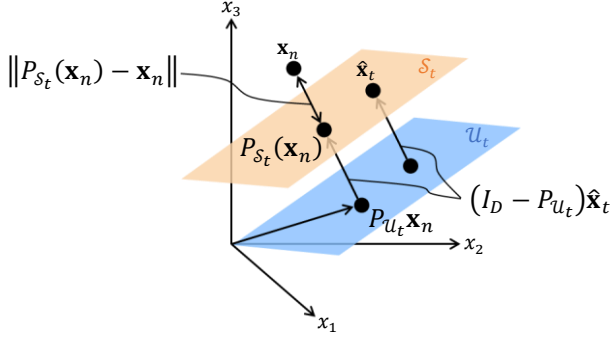

Fig. 2. Geometric meaning of (6). $\mathcal{U}_t$ component of $P_{S_t}(\mathbf{x}_n)$ and its orthogonal complement in case $D = 3, d_t = 2$. The $\|P_{S_t}(\mathbf{x}_n) - \mathbf{x}_n\|$ in (5) represents the orthogonal projection length from $\mathbf{x}_n$ to $S_t$.

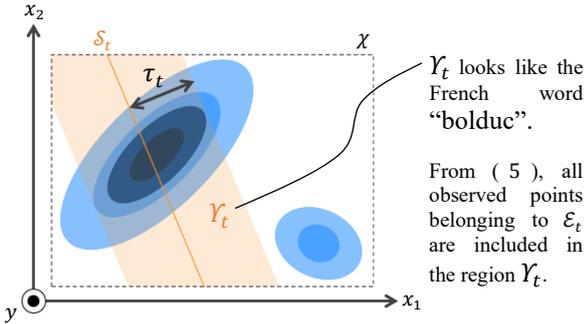

Fig. 3. An example of region $\Upsilon_t$ covering LSoD $\mathcal{E}_t$ when $D = 2, d_t = 1$. The region $\Upsilon_t$, which characterizes our method, looks like the French word "bolduc", which in English means gift ribbon or tape.

holds. $k(P_{S_t}(\mathbf{x}_n), \mathbf{x}_n) \geq \lambda_t$ and $\|P_{S_t}(\mathbf{x}_n) - \mathbf{x}_n\| \leq \tau_t$ are equivalent with the threshold $\tau_t$ corresponding to $\lambda_t$, since the aforementioned kernel function was adopted. ∎

The kernel function adopted in Theorem 1 includes, for example, the squared exponential (SE) kernel, $k(\mathbf{x}_i, \mathbf{x}_j) = \theta_\sigma^2 \exp\left(-\frac{1}{2}\frac{r^2}{\theta_l^2}\right)$, and the Matérn 5/2 kernel, $k(\mathbf{x}_i, \mathbf{x}_j) = \theta_\sigma^2 \left(1 + \frac{\sqrt{5}r}{\theta_l} + \frac{5r^2}{3\theta_l^2}\right)\exp\left(-\frac{\sqrt{5}r}{\theta_l}\right)$, where $r = \|\mathbf{x}_i - \mathbf{x}_j\|$, $\theta_\sigma$ and $\theta_l$ represent signal standard deviation and length scale, respectively. Figure 2 shows the geometric meaning of (6) when $D = 3, d_t = 2$. $P_{\mathcal{U}_t}\mathbf{x}_n$ and $(I_D - P_{\mathcal{U}_t})\hat{\mathbf{x}}_t$ are the $\mathcal{U}_t$ component of $P_{S_t}(\mathbf{x}_n)$ and its orthogonal complement, respectively. The $\|P_{S_t}(\mathbf{x}_n) - \mathbf{x}_n\|$ in (5) represents the orthogonal projection length from $\mathbf{x}_n$ to $S_t$. Therefore, all observed points belonging to $\mathcal{E}_t$ are included in the region $\Upsilon_t$ whose Euclidean distance from $S_t$ is less than or equal to $\tau_t$. An example of region $\Upsilon_t$ when $D = 2$ and $d_t = 1$ is shown in the shaded area in Fig. 3. From the above, Theorem 1 means that $\mathcal{E}_t$ in (3) can be calculated from the projection length without calculating exponents included in the SE kernel or the Matérn 5/2 kernel by using the assumption of (4). Note that if the ARD kernel is adopted, the points belonging to $\mathcal{D}_t$ are standardized with the length scale vector estimated when the LGPR model was built from $\mathcal{E}_{t-1}$ before extracting $\mathcal{E}_t$.

**Algorithm 3** BOLDUC
1: Initialize data: $\mathcal{D}_{N_0} = \{(\mathbf{x}_n, y_n)\}_{n=1,2,\ldots,N_0}$.
2: **for** $t = N_0, N_0 + 1, \ldots, N - 1$ **do**
3:    Define low-dimensional search space $S_t$.
4:    Extract LSoD $\mathcal{E}_t$ from $\mathcal{D}_t$ using (3), (5) or strategies described in Subsect. III D–F.
5:    Estimate kernel hyperparameter $\boldsymbol{\theta}(\mathcal{E}_t)$ and train LGPR model $\mathcal{M}(\mathcal{E}_t)$.
6:    Suggest $\mathbf{x}_{t+1} = \arg\max_{\mathbf{x} \in (S_t \cap \chi)} \alpha(\mathbf{x}|\mathcal{E}_t)$ using $\mathcal{M}(\mathcal{E}_t)$.
7:    Observe $y_{t+1} = f(\mathbf{x}_{t+1}) + \epsilon_{t+1}$.
8:    Augment the data $\mathcal{D}_{t+1} = \mathcal{D}_t \cup \{(\mathbf{x}_{t+1}, y_{t+1})\}$.
9: **end for**
10: Output $\hat{\mathbf{x}}_N = \mathbf{x}_b$, where $b = \arg\min_{n=1,2,\ldots,N} y_n$.

### C. BOLDUC Procedure

Algorithm 3 shows the BOLDUC procedure. Lines 4–6 differ from Algorithm 2, where $\mathcal{E}_t$ is used instead of $\mathcal{D}_t$. Regarding line 4, three strategies for extracting the LSoD $\mathcal{E}_t$ without directly specifying $\lambda_t$ in (3) are described in Subsect. III D–F. Each can be converted into the form of (3) or (5).

$\mathcal{E}_t$ does not contain samples with negligible contributions to the prediction of points in $(S_t \cap \chi)$. The prediction accuracy in $(S_t \cap \chi)$ can be expected to be equal to or better than global GPR using $\mathcal{D}_t$, since length scales suitable for representing the local structure of $f$ in $(S_t \cap \chi)$ are estimated at line 5. The time complexity of matrix inversion is reduced from $\mathcal{O}(|\mathcal{D}_t|^3)$ to $\mathcal{O}(|\mathcal{E}_t|^3)$. BOLDUC inherits the theoretical guarantees in [14] when LineBO is adopted as BOLD because LGPR does not affect those guarantees.

### D. LSoD extraction strategy 1

It is preferable that the GPR calculation time is short with respect to the observation time to obtain as many samples as possible in a limited period. A strategy to control that time is most practical in time-critical situations.

We generate LSoD $\mathcal{E}_t$ by extracting from $\mathcal{D}_t$ the top $M_t$ samples with the highest contribution to $(S_t \cap \chi)$ to control the time complexity. If $M_t \geq |\mathcal{D}_t|$, then $\mathcal{E}_t = \mathcal{D}_t$. Here, $M_t$ is the number of elements of $\mathcal{E}_t$ specified by the user. $M_t$ needs to be set sufficiently large to maintain LGPR prediction accuracy in $d_t$-dimensional space, but as long as $d_t$ is set small, such as 3 or less, its value will be sufficiently small compared with several thousand, for which the calculation time of an inverse matrix becomes unrealistic. When the $M_t$ is a constant $M$ independent of $t$, adaptive GPR can be realized in which the structure of $f$ is at first coarsely and then finely captured as time $t$ progresses. This is because GPR captures the global structure from $\mathcal{D}_t$ in the early stage, when $|\mathcal{D}_t| \leq M$ holds, and LGPR captures the local structure from $\mathcal{E}_t$ in the later stage. Although the time complexity of sorting $\mathcal{O}(|\mathcal{D}_t|\log|\mathcal{D}_t|)$ is added, it is smaller than the effect of reducing the calculation time for the inverse matrix.

Adopting the kernel function of Theorem 1 extracts $M_t$ samples with small $\|P_{S_t}(\mathbf{x}_n) - \mathbf{x}_n\|$ in (5).

*E. LSoD extraction strategy 2*

At the stage when the user has specified the domain $\chi$ based on experience and preliminary experiments, the objective $f$ within $\chi$ may have a shape that is not too smooth and does not oscillate excessively. When a kernel with higher similarity is adopted as the Euclidean distance is shorter, the user can specify $\tau_t$ in (5). If it is desired to at first coarsely and later finely control the search as time $t$ progresses, $\tau_t$ should be gradually made smaller.

*F. LSoD extraction strategy 3*

From the standpoint of regarding LGPR as a local approximation of global GPR, there are cases where we want to control or visualize the approximation accuracy. Although it does not fit the hypothesis that the prediction accuracy of LGPR can be expected to be equal to or higher than that of global GPR because $\boldsymbol{\theta}(\mathcal{E}_{t-1})$ is better suited than $\boldsymbol{\theta}(\mathcal{D}_t)$ to represent the local structure of $f$, we assume $\boldsymbol{\theta}(\mathcal{D}_t) \approx \boldsymbol{\theta}(\mathcal{E}_{t-1})$ to discuss the approximation framework.

In this case, first, the threshold value $C_t$ regarding the cumulative contribution rate in the range $[0,1]$ is received from the user. Next, the $t$ observed points included in $\mathcal{D}_t$ are sorted in descending order of contribution $\max_{\mathbf{x} \in (\mathcal{S}_t \cap \chi)} k(\mathbf{x}, \mathbf{x}_n)$ to $(\mathcal{S}_t \cap \chi)$. Let $\eta = 1, 2, \ldots, t$ denote their index after sorting. Then, let $M_t$ be the minimum $h (\geq 1)$ at which the cumulative contribution rate

$$c_t(h) = \sum_{\eta=1}^{h} \max_{\mathbf{x} \in (\mathcal{S}_t \cap \chi)} k(\mathbf{x}, \mathbf{x}_\eta) \bigg/ \sum_{\eta=1}^{t} \max_{\mathbf{x} \in (\mathcal{S}_t \cap \chi)} k(\mathbf{x}, \mathbf{x}_\eta)$$

is equal to or greater than the threshold $C_t$. Finally, $\mathcal{E}_t$ is extracted as $M_t$ samples from $\mathcal{D}_t$ in ascending order of index $\eta$ after the sorting:

$$\mathcal{E}_t = \{(\mathbf{x}_\eta, y_\eta) | \eta = 1, 2, \ldots, M_t\}.$$

As the threshold $C_t$ increases, $|\mathcal{E}_t|$ also increases, so the approximation accuracy in $(\mathcal{S}_t \cap \chi)$ increases. Although the user cannot grasp $\mathcal{O}(|\mathcal{E}_t|^3)$ in advance, the approximation accuracy can be controlled by specifying $C_t$ in a manner similar to principal component analysis.

## IV. EXPERIMENTS

*A. Evaluations using benchmark functions*

We evaluated BOLDUC using two benchmark functions as the objective function. One was the 20D Ackley function,

$$f(\mathbf{x}) = 20 + e - 20 \exp\left(-0.2\sqrt{\frac{1}{20}\sum_{d=1}^{20} x_d^2}\right) +$$

$\exp\left(\frac{1}{20}\sum_{d=1}^{20} \cos 2\pi x_d\right)$, $x_d \in [-32.768, 32.768]$, for all $d = 1, 2, \ldots, 20$. The other was the 20D Rosenbrock function, $f(\mathbf{x}) = \sum_{d=1}^{20-1}(100(x_{d+1} - x_d^2)^2 + (x_d - 1)^2)$, $x_d \in [-5, 10]$, for all $d = 1, 2, \ldots, 20$. The minimum value $f(\mathbf{x}_{opt})$ is 0 in both. Figure 4 shows the shape of each function in the two-dimensional case for reference.

We adopted CoordinateLineBO (CLBO) [14] as BOLD to facilitate replication experiments and interpretations of evaluation results. In this section, we refer to CLBO utilizing LGPR as 1D-BOLDUC. We normalized the domain $\chi$ to $[-0.5, 0.5]^{20}$ in CLBO. At line 3 in Algorithm 2, the one-dimensional search space $\mathcal{S}_t$ was updated every $5d_t (= 5)$ advances of $t$. The direction of the line at the time of switching was determined cyclically in a predetermined order, but not randomly from the D coordinate axis directions. We used a SE kernel. The acquisition function was LCB with $\kappa_t = 2$. $N_0$ and $N$ were $D$ and 1000, respectively. We used the approximation expressed in (4) for 1D-BOLDUC. We compared the evaluation with the 20D Ackley function with CMA-ES [33], standard BO, TuRBO-1 [24] (which focuses on local regions like our BOLDUC), and CLBO. In the case of the 20D Rosenbrock function, we added REMBO [18] with a two-dimensional embedding space to the comparison methods. The comparison methods did not include REMBO in the Ackley case because the embedded two-dimensional space accidentally included the optimal point $\mathbf{x}_{opt}$, which is the origin, so the comparison lacked fairness. As the results varied depending on the initial data $\mathcal{D}_{N_0}$, each function was optimized with thirty different initial data for each method. The same $\mathcal{D}_{N_0}$ was given except for CMA-ES and REMBO.

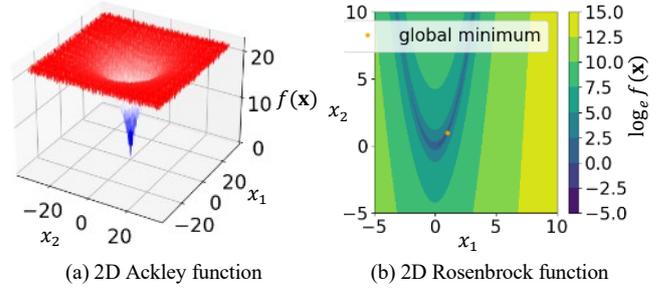

(a) 2D Ackley function  (b) 2D Rosenbrock function

Fig. 4. Visualization of the benchmark functions in the 2D cases for reference. (a) The Ackley function has relatively weak multimodality and locality around the global minimum. (b) The Rosenbrock function has strong dependence between neighboring parameters.

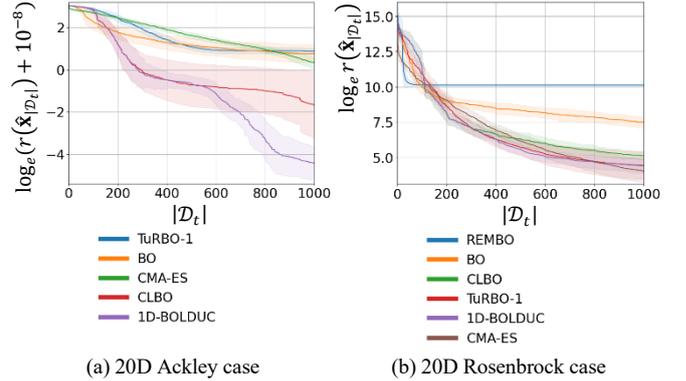

(a) 20D Ackley case  (b) 20D Rosenbrock case

Fig. 5. Logarithmic regrets for the evaluation of 1D-BOLDUC (LSoD extraction Strategy 1). Each solid line and shaded area represent the mean and standard deviation of 30 trials. The method names in the legend are arranged in descending order of the means at $|\mathcal{D}_t| = 1000$.

Figure 5 shows the thirty-time averages and standard deviations of the logarithmic regrets of 1D-BOLDUC with $M = 200$ in Strategy 1 and the compared methods. In (a) the Ackley case, a decimal $10^{-8}$ was added to prevent logarithmic divergence. The method names in the legend are arranged in descending order of their average values. In (a) the Ackley case, the regret of the 1D-BOLDUC was the smallest. In (b) the Rosenbrock case, the regret $r(\hat{\mathbf{x}}_{1000})$ of 1D-BOLDUC was the second smallest after CMA-ES, and the difference between the two was not large. REMBO had larger regret because the assumption about the effective dimension was invalid.

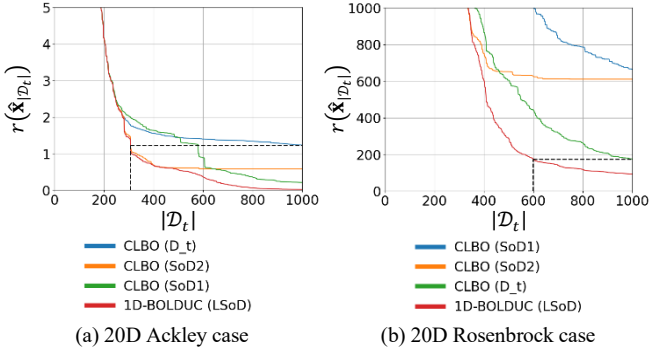

(a) 20D Ackley case      (b) 20D Rosenbrock case

Fig. 6. Simple regrets for the evaluation of SoD extraction methods. Each solid line and shaded area represent the mean of 30 trials. The method names in the legend are arranged in descending order of the means at $|\mathcal{D}_t| = 1000$.

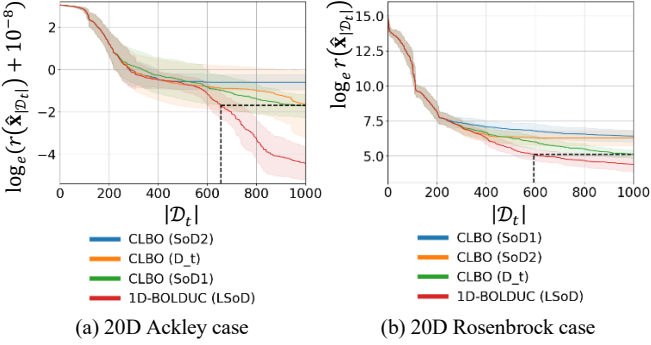

(a) 20D Ackley case      (b) 20D Rosenbrock case

Fig. 7. Logarithmic regrets for the evaluation of SoD extraction methods. Each solid line and shaded area represent the mean and standard deviation of 30 trials. The method names in the legend are arranged in descending order of the means at $|\mathcal{D}_t| = 1000$.

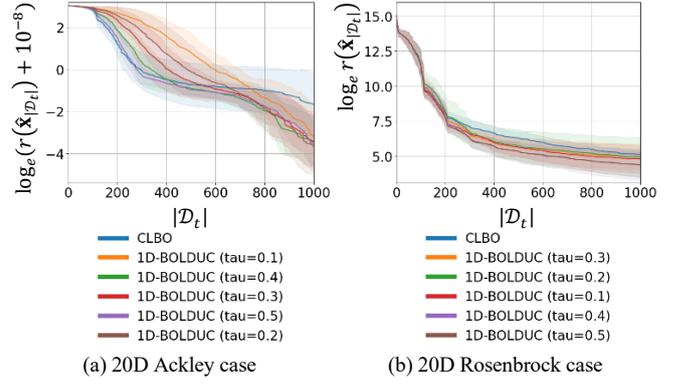

(a) 20D Ackley case      (b) 20D Rosenbrock case

Fig. 8. Logarithmic regrets for the evaluation of 1D-BOLDUC (LSoD extraction Strategy 2). Each solid line and shaded area represent the mean and standard deviation of 30 trials. The method names in the legend are arranged in descending order of the means at $|\mathcal{D}_t| = 1000$.

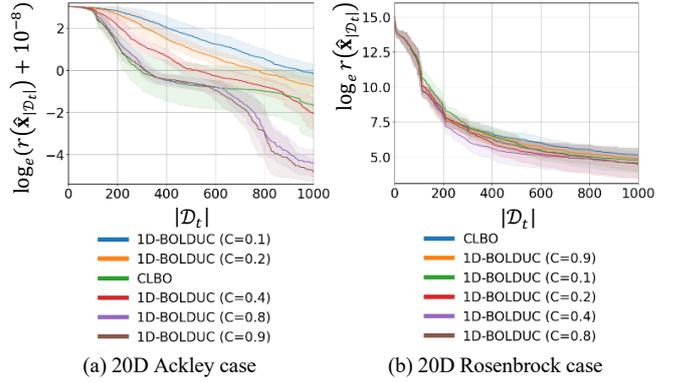

(a) 20D Ackley case      (b) 20D Rosenbrock case

Fig. 9. Logarithmic regrets for the evaluation of 1D-BOLDUC (LSoD extraction Strategy 3). Each solid line and shaded area represent the mean and standard deviation of 30 trials. The method names in the legend are arranged in descending order of the means at $|\mathcal{D}_t| = 1000$.

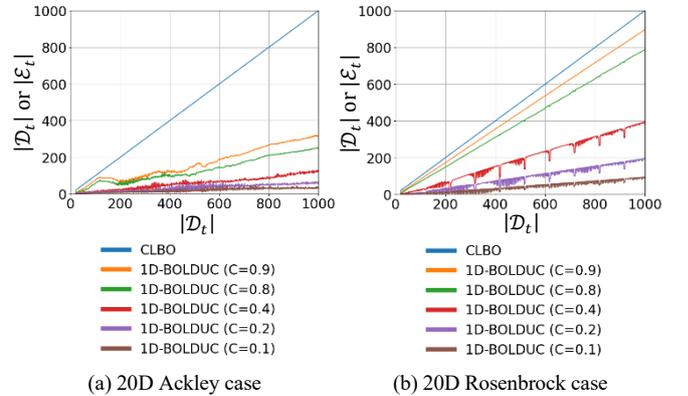

(a) 20D Ackley case      (b) 20D Rosenbrock case

Fig. 10. Numbers of samples $|\mathcal{D}_t|$ and $|\mathcal{E}_t|$ extracted for each $C_t$ (averages of 30 trials).

To evaluate the effectiveness of our LSoD extraction method, we compared four CLBOs that changed the data used in GPR. The four data types were $\mathcal{D}_t$, LSoD $\mathcal{E}_t$ with $M = 200$ in Strategy 1, SoD1 with 200 samples randomly selected from $\mathcal{D}_t$, and SoD2 with 200 samples extracted in ascending order of observed values included in $\mathcal{D}_t$. Each function was optimized thirty times by each method, and Fig. 6 shows the simple regrets of each. We obtained the simple regret by CLBO at $|\mathcal{D}_t| = 1000$ by 1D-BOLDUC based on $\mathcal{E}_t$ at $|\mathcal{D}_t| = 307$ and 600 in cases (a) and (b), respectively, so the respective search improvement rates were about 69% and 40%. Figure 7 shows the thirty-time averages of the logarithmic regrets of each method, which showed respective improvement rates of about 35% and 41%. In both cases (a) and (b), the search efficiency was better when $\mathcal{E}_t$ was used than when SoD1 and 2 were used. This confirmed an improvement effect on search efficiencies by LSoD extraction according to contribution to $\mathcal{S}_t$.

Figure 8 shows the transitions of the logarithmic regrets of CLBO and 1D-BOLDUC with $\tau_t = 0.1, 0.2, 0.3, 0.4, 0.5$ in the normalized domain in Strategy 2. In both cases (a) and (b), 1D-BOLDUC had smaller regrets than did CLBO, regardless of the value of $\tau_t$.

Figure 9 shows the logarithmic regrets of CLBO and 1D-BOLDUC with $C_t = 0.1, 0.2, 0.4, 0.8, 0.9$ in Strategy 3. In both cases (a) and (b), the search efficiency of 1D-BOLDUC with $C_t \geq 0.4$ was higher than that of CLBO. Since the difference was small when $C_t = 0.8$ and $0.9$, we recommend this level of $C_t$ value.

Figure 10 shows averages of numbers of samples $|\mathcal{D}_t|$ and $|\mathcal{E}_t|$ extracted for each $C_t$. At $C_t = 0.1, 0.2$ in (a), the averages of $|\mathcal{E}_t|$ were less than 100 at maximum. In the case of $C_t = 0.1, 0.2$, $|\mathcal{E}_t|$ was too small to guarantee GPR prediction accuracy in the low-dimensional search space, and the search efficiency shown in Fig. 9a was considered to be worse than CLBO. At each $C_t$ in (b), $|\mathcal{D}_t|$ and $|\mathcal{E}_t|$ were approximately proportional. Since the locality of the Ackley function is high, $|\mathcal{E}_t|$ in (a) was smaller than that in (b) when comparing with the same $C_t$ and $|\mathcal{D}_t|$.

Figure 11 shows the averages of the length scale $\theta_l(\mathcal{D}_t)$

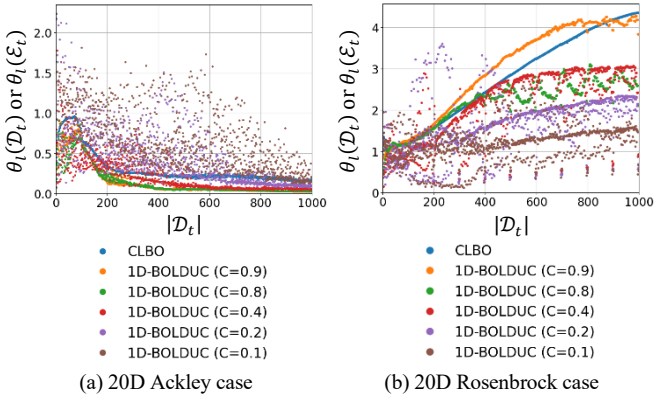

(a) 20D Ackley case  (b) 20D Rosenbrock case

Fig. 11. Length scales $\theta_l(\mathcal{E}_t), \theta_l(\mathcal{D}_t)$ estimated in each method (averages of 30 trials).

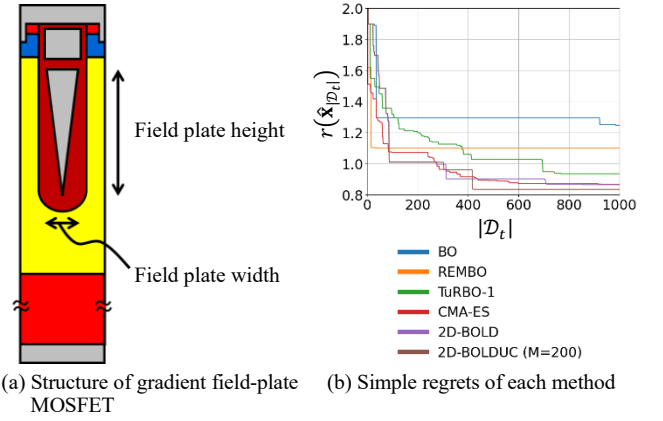

(a) Structure of gradient field-plate MOSFET  (b) Simple regrets of each method

Fig. 12. Results from the automatic design task. In (b), The method names in the legend are arranged in descending order of the simple regrets.

estimated by global GPR in CLBO and the length scale $\theta_l(\mathcal{E}_t)$ estimated by LGPR for each $C_t$ in 1D-BOLDUC. These were the values in normalized domain. Since the locality of the Ackley function is high, the length scale in (a) was smaller than that in (b), under the same conditions. In (a), the length scale was also larger than CLBO at $C_t = 0.1, 0.2$, where the regret was larger than in CLBO in Fig. 9a. For $C_t = 0.4, 0.8, 0.9$, where the regret was smaller than CLBO, the length scale was also smaller than in CLBO. If $C_t$ is too small, the observation points included in $\mathcal{E}_t$ cannot cover the low-dimensional search region $(\mathcal{S}_t \cap \chi)$, so the length scale that can reproduce the local structure of the objective $f$ cannot be estimated. However, if $C_t$ is within a certain range, a length scale capable of finely reproducing the local structure is estimated. In Fig. 11b, the smaller $C_t$ is, the smaller the length scale. In Fig. 9b, for any $C_t$ less than 1, the regrets were smaller than in CLBO corresponding to $C_t = 1$, so we can see that the structure in $(\mathcal{S}_t \cap \chi)$ of the Rosenbrock function was reproduced when $C_t < 1$.

### B. Automatic design of a power semiconductor device

We evaluated the effectiveness of BOLDUC by applying it to a design task for a power semiconductor device. We constructed an automatic design environment by combining our optimization method and technology computer-aided design (TCAD), which simulates characteristic values according to design variables on a computer. The design target was a MOSFET structure with a gradient field plate, described in [10] and shown in Fig. 12a. The design target **x** input to TCAD was a vector consisting of six design variables, including the field plate width and height. The breakdown voltage $V_B$ and the specific on-resistance $R_{ON}A$ were output from TCAD. $V_B$ must be greater than 110V to prevent device breakage. The smaller the $R_{ON}A$, the less the power loss, reducing carbon dioxide emissions. The objective function to be minimized was defined as

$$f(\mathbf{x}) = R_{ON}A/30 + 10\,\text{ReLU}(110 - V_B),$$

as in [10], where ReLU is a rectified linear unit function.

In BOLD, we fixed the dimension $d_t$ of $\mathcal{S}_t$ to 2 regardless of time $t$. We set one of the two basis vectors of the two-dimensional linear subspace $\mathcal{U}_t$ as the cyclically selected coordinate axis and determined the other randomly, not necessarily parallel to the coordinate axes. We updated $\mathcal{S}_t$ every $5d_t (= 10)$ advances of $t$. We thus refer to this method as 2D-BOLD. We did not adopt CLBO [14] as BOLD because like the coordinate descent method it can, in principle, fall into a local minimum. We adopted LSoD extraction Strategy 1 for LGPR and set $M$ to 200, which was reasonable to cover the two-dimensional $\mathcal{S}_t$ by the observation points included in $\mathcal{E}_t$. We performed comparisons with 2D-BOLDUC, 2D-BOLD, CMA-ES [33], standard BO, REMBO [18], and TuRBO-1 [24]. Due to a limited budget for TCAD simulations, we optimized each method once with $N = 1000$. Other than CMA-ES and REMBO, we set $N_0 = D (= 6)$ and used the same initial points $\{\mathbf{x}_n\}_{n=1,2,\ldots,N_0}$ in the Sobol sequence.

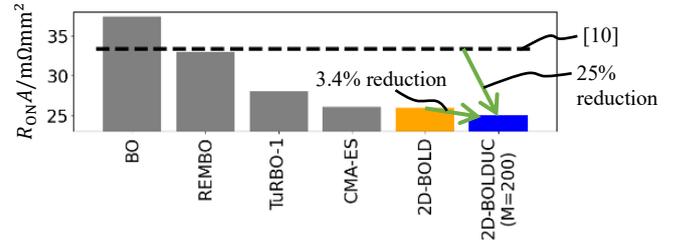

Fig. 13. Obtained minimum specific on-resistances by each method.

Figures 12b and 13 show transitions of the simple regrets and the obtained minimum $R_{ON}A$s under the condition that $V_B$ is 110 or more in each method. Although the parameter space dimensionality was 6, which is not high, standard BO was inferior to other methods. Our 2D-BOLDUC obtained the smallest objective function value and $R_{ON}A$ compared to the other methods. $R_{ON}A$ was approximately 25% smaller than CLBO in [10] and 3.4% better than without LGPR.

## V. CONCLUSION

We proposed a novel BO algorithm, BOLDUC, to scale BO to higher dimensions. Focusing on the fact that BOLD requires prediction only in low-dimensional search regions, we presented a LSoD extraction framework to realize LGPR specialized to those regions.

By LGPR, calculation times for matrix inversion can be suppressed to within a certain time, so it works even with an increased number of observation points. We estimated the length scale that models the local structure of the objective function from LSoD and, using benchmark functions, we confirmed an improvement in search efficiency as compared to BOLD and other methods. Constructing an automatic design environment for a power semiconductor device, our method successfully found a design vector value that yielded better $R_{ON}A$ than [10], BOLD, and the other compared methods.

ACKNOWLEDGMENT

Thanks to T. Inokuchi and K. Nakata for giving us the opportunity to team up. The authors are grateful to S. Shioda and T. Ohashi for assistance with setting up the TCAD simulations and thank T. Nishiwaki, T. Kachi, and H. Nakagawa for valuable discussions and feedbacks.